%% file: main.tex
\documentclass[twoside,11pt,abbrvbib]{article}

\usepackage{jmlr2e}

\usepackage[T1]{fontenc}
\usepackage{booktabs}
\PassOptionsToPackage{hyphens}{url}
\usepackage{url}

\usepackage{bbm}
\usepackage{tikz}
\usepackage{lipsum}
\usepackage{array}
\usepackage{booktabs,multicol,multirow}
\usepackage{xcolor,colortbl}

\newcommand{\ie}[1]{\textit{i.e.}}
\newcommand{\eg}[1]{\textit{e.g.}}

\ShortHeadings{Span-Oriented Information Extraction}{Ding, Yankoski and Weninger }
\firstpageno{1}

\begin{document}

\title{Span-Oriented Information Extraction \\ \normalsize{A Unifying Perspective on Information Extraction}}

\author{\name Yifan Ding \email yding4@nd.edu \\
       \addr University of Notre Dame\\
       Notre Dame, IN, USA
       \AND
       \name Michael Yankoski \email myankosk@colby.edu \\
       \addr Colby College\\
       Waterville, ME, USA
       \AND
       \name Tim Weninger \email tweninger@nd.edu \\
       \addr University of Notre Dame\\
       Notre Dame, IN, USA}

\editor{}

\maketitle

\begin{abstract}
Information Extraction refers to a collection of tasks within Natural Language Processing (NLP) that  identifies sub-sequences within text and their labels. These tasks have been used for many years to link extract relevant information and to link free text to structured data. However, the heterogeneity among information extraction tasks impedes progress in this area. We therefore offer a unifying perspective centered on what we define to be \textit{spans} in text. We then re-orient these seemingly incongruous tasks into this unified perspective and then re-present the wide assortment of information extraction tasks as variants of the same basic Span-Oriented Information Extraction task. 

\end{abstract}


\section{Introduction}

Most of the unstructured or semi-structured text data available today is available as a sequence-of-bytes (\ie, a sequence of text characters). From these sequences-of-bytes, both humans and machines create whole-words and sentences, and then more-complex paragraphs and then stories that, at their limit, encompass complex narratives, emotions, and literature. Without extra effort, these sequences-of-bytes cannot be effectively used to extract insights or create meaning. Indeed, most natural language processing (NLP) systems, including large language models (LLMs), do \textit{not} take as input data in its native sequence-of-bytes format. Instead, raw characters are first broken up into individual tokens, which are typically individual words, but can also be sub-word chunks, that are gleaned from the sequence-of-bytes. These tokens are what is actually used to train NLP systems, So when an NLP system translates text or answers a question, it is actually generating sequences-of-tokens, not individual characters or bytes. 

Contrary to the token-by-token view used by most NLP systems, real-world entities and objects are often represented by multi-token sub-sequences (\ie, names) where a span of one or more tokens represents a single entity. The NLP task that aims to identify pertinent sub-sequences and provide them a label or link them to some external structured knowledge is called \textit{information extraction}. Because the surface text representing an entity is often comprised of more than one token, these tokens are commonly referred to as a \textit{span}.



\begin{figure}[t]
    \centering
    \includegraphics[width=1\linewidth]{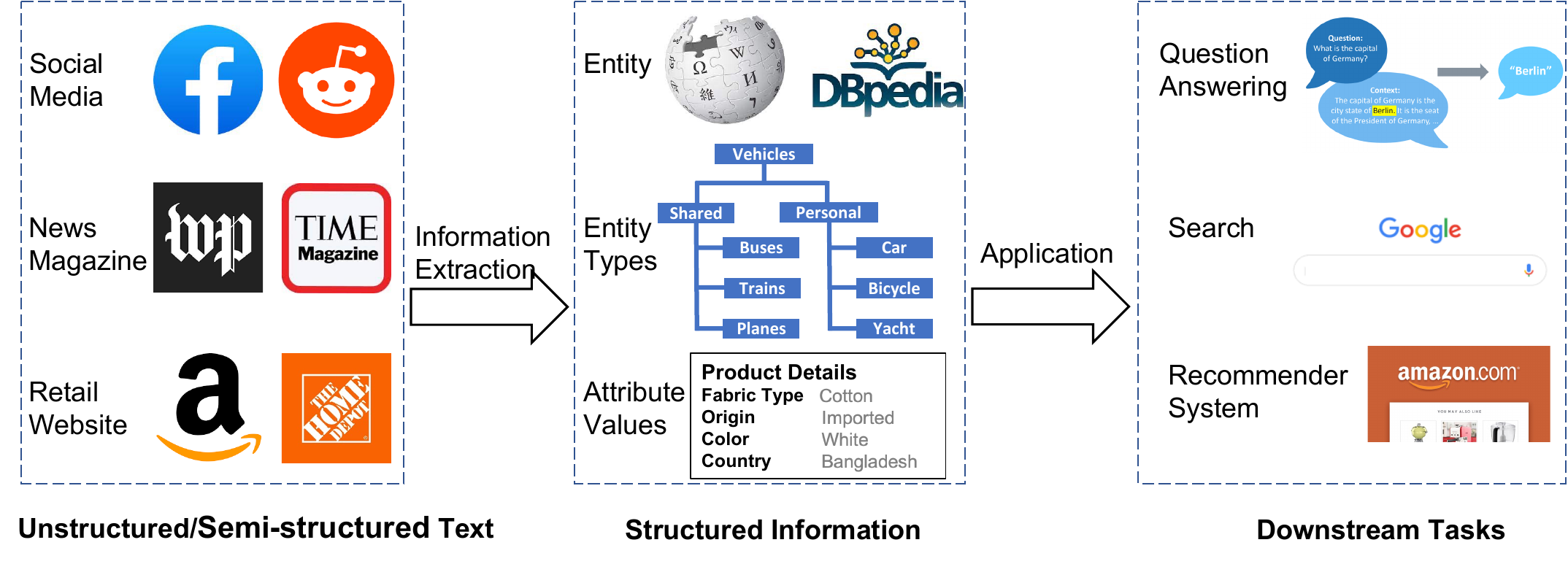}
    \caption{An overview of information extraction tasks. The goal of information extraction is to identify sub-sequences within unstructured or semi-structured text information and link them to certain class-labels, entities in a knowledge base, or other items within some structured database. This structured information plays a central role in many downstream applications such as question answering and recommender systems.}
    \label{fig:application}
\end{figure}

Take, for example, the entity \textsf{Tim Cook}, the CEO of Apple Inc., who frequently appears in news articles and social media posts. If we could map the surface form (\ie, the sequence-of-tokens) of \texttt{Tim Cook} to some structured knowledge listing in, say, Wikipedia (\url{wiki/Tim_Cook}), then all of the related information in that knowledge base could be leveraged to inform other downstream tasks, like mitigating citation hallucination in LLMs~\cite{Mikolov-NIPS'13-word2vec}, Wikification~\cite{Ferragina-CIKM'10-tagme}, question answering~\cite{Rajpurkar-EMNLP'16-squad}, and an overall better understanding of the text as illustrated in Fig.~\ref{fig:application}. But, all of these potential advances are first predicated on how we represent spans within the NLP system.

Unlike the tokenization task and the related tasks of parsing, chunking, lemmatization, and stemming, each having an enormous literature surrounding them, the representation of spans in information extraction tasks is relatively unprincipled. Mostly by coincidence, it appears that a handful of span-oriented representations have emerged in the information extraction literature. Despite their centrality in many NLP systems, these mostly ad hoc methodologies have not yet been cataloged; yet they constitute a critical task in NLP.

Most information extraction systems either (1) find relevant sub-sequences and then use them to predict a label or (2) have a label and then attempts to find an appropriate sub-sequence that represents that label. But these one-way transformations can be problematic because they do not consider the duality that exists between sub-sequences and their labels. This is one reason why, for example, zero-shot entity linking remains difficult: the one-way entity linking transformation that links known-entities to their surface forms cannot be used if the entity has never been seen~\cite{sevgili-SW'22-survey, Li-TKDE'21'-NER_Survey}.

In contrast, we view a span as simultaneously possessing (1) a sub-sequence of tokens representing the entity (\ie, a surface form) and (2) a meaning/label. This is natural because, when reading, humans simultaneously derive meaning from text while also using that meaning to understand the surface forms that encode those concepts. With this view, many challenges in information extraction could be re-considered with respect to this duality. The rise of LLMs and prompt tuning also fit nicely into this paradigm because the emergent ability of LLMs (and their faux reasoning ability) could be further empowered if the generated tokens could consider the duality inherent in their spans.

By re-orienting the disparate representations used across the enormous variety of tasks in NLP around a unified definition of a span, we show that various information extraction tasks are actually not that different after all. We call this perspective \textit{Span-Oriented Information Extraction}. As the name implies, the goal of span-oriented information extraction is to (1) extract spans from sequences-of-tokens and (2) assign each span to a well-defined entity. Traditional information extraction systems typically ignore the first step and focus exclusively on the second,
but, as we shall see, the way in which spans are identified and represented has critical implications on the overall system.

\subsection{A Brief History of Spans in Information Extraction}


The use of spans in information extraction dates back to the early days of Computational Linguistics and NLP~\cite{Cowie-Comunications_of_ACM'96-IE}. In its early forms, spans were the basic units used in parsing and part of speech chunking tasks. The Linguistic String Project (LSP)~\cite{Transporting-TOIS'85-Marsh} at New York University was one of the earliest information extraction research and development projects. LSP began with a parsing program to obtain the syntactic relations among words in a sentence. The first implementation of the LSP parser and grammar was introduced in 1967 with two key components:  a set of formal rewriting rules, and a set of restrictions. The applications of the LSP system drew upon a so-called \textit{sublanguage grammar}, where documents could be parsed with quasi-grammatical rules and further mapped into an early incarnation of a span, called an \textit{information format}. LSP mainly focused on medical data analysis including medical test reports, hospital discharge records, and patient diagnosis history, but its impact can still be seen in contemporary NLP tools. 

Beginning in the late 1980s, the Message Understanding Conference (MUC) advanced the development of information extraction and established this task as one of the key branches of NLP. The Defense Advanced Research Projects Agency (DARPA) funded MUC also introduced early evaluation processes for information extraction systems \cite{Chinchor-MUC'93-muc_5, Grishman-COLING'96-MUC-6}. Each team attending the MUC conference was required to submit an information extraction system for the given task. MUC-1 to MUC-4 focused on \textit{template filling} of naval military intelligence and terrorist incidents, with the goal to fill-out pre-identified spans within text-templates. Evaluation metrics used in information retrieval such as precision and recall were introduced formally in MUC-3. In the next iteration, MUC-5 expanded beyond military domains to joint ventures in the financial sector and in microelectronics~\cite{Chinchor-MUC'93-muc_5}. 
MUC-6 introduced the concept of a \textit{named entity} and formalized the task of named entity recognition (NER), with its corresponding evaluation metrics~\cite{Grishman-COLING'96-MUC-6}. The goal of NER, as originally defined, was to identify the spans within text that correspond to people, organizations, and geographic locations. With the intervening development of knowledge bases like Wikipedia~\cite{Wikipedia-HC}, information extraction tasks were refined to not just identify NER-spans, but to link them to their corresponding entry in the knowledge base. The development of digital encyclopedias was coincident with MUC-7, which introduced the task of relation extraction and paved the way for knowledge bases and knowledge graphs as we know them today.

The Automatic Content Extraction (ACE) program began shortly after MUC was concluded.
ACE mainly focused on news and conversational data including broadcast conversations, newspapers, and conversational telephone speech. 
It's goal was entity recognition and labeling on a large natural language corpus. In pursuit of that goal, ACE also established the use of spans and now-standard labels used for named entities and relation types. 

Since 2007, NIST has organized the Text Analysis Conferences (TAC), which has been another driving force in NLP research. Like MUC, TAC had a task submission and evaluation setup, and then proceedings to discuss submitted methods and models. Many well-known information extraction tasks, including entity linking, slot filling, fine-grained entity typing and others have been conceived and advanced by participating teams working on these challenges and datasets. 

These initial competitions paved the way for the myriad of information extract tasks, which are used to understand, generate, and reason about text-based natural language. Despite the variety of information extraction tasks and the innumerable ways that spans are represented in these tasks, there does not appear to be a coherent view of the definition and role of a span within the various tasks in information extraction. We argue in the present work that a span is a foundational building block which, if carefully constructed and considered, naturally ties together these tasks.

\subsection{Formal Definition of a Span}


Formally, given a document $d$ containing a sequence of bytes (typically characters) that are grouped into tokens $d = \langle t_1, t_2, \ldots, t_{k=\ell(d)}\rangle$ where $\ell(d)$ is the length of $d$, a \textit{span} is a triple $s=(b, e, c)$ with a beginning token-index $b$, ending token-index $e$, and class label(s) $c$. 

Informally, a span represents a duality comprised of: (1) a surface form (\ie, a subset of the original text), and (2) one or more corresponding span classes (\ie, the class label(s)). Thus the goal of the span-oriented information extraction task is to extract the correct spans (if not given) with correct boundaries and assign correct class labels. The makeup of the class set $C$ defines two varieties of information extraction: (1) closed-world, which seeks to assign each span to one or more classes selected from a pre-defined class set $C$, and (2) open-world, which does not contain a pre-defined class set, but rather may assign labels from any source. This definition is operationalized by most popular software packages like Spacy~\cite{Honnibal-17-spacy} and Stanford's CoreNLP~\cite{Manning-ACL'14-stanford_core_nlp}, which both implement a \textit{span} as a built-in class. 

\subsection{Road Map}
With the span definition in hand, the goal of the present work is to re-imagine the landscape of the varieties of span-oriented information extraction tasks along three dimensions: (1) information extraction \textit{tasks}, (2) information extraction \textit{evaluation}, and (3) information extraction \textit{models}.

\paragraph{Information Extraction Tasks} Information extraction tasks appear to vary significantly in their inputs and outputs. For example, named entity recognition seeks to label named entities (\eg person, place, organization, or other proper nouns) within sentences; and  machine reading comprehension tasks seeks to answer questions based on information in the background context. These two tasks appear to be quite different; however, if we re-imagine these tasks from the perspective of the span, different information extraction tasks are naturally aligned. As we shall see in Section 2, vastly different information extraction tasks can be re-imagined so as to generate the same span outputs with only slightly different inputs.

\paragraph{Information Extraction Evaluation} Because of the varieties of information extraction tasks, there is no shortage of ways to evaluate the performance of various algorithms and models. It is important to encourage broad evaluation of these tasks; however, a span-oriented consideration of information extraction tasks permits, as we shall see in Section 3, a natural comparison of the predictions made by different tasks.

\paragraph{Information Extraction Models}
If we indeed re-imagine information extraction tasks as having the same span-based outputs, and similar evaluation regimes, then, as we shall see in Section 4, it becomes natural to consider the various algorithms and models as simple variants of one another.

The present work endeavors to present the wide assortment of information extraction tasks as variants of the same basic span-prediction algorithm. With this in mind, it is possible that algorithms, models, architectures, and intuitions developed in the solution to one task might be used to solve others.


\section{Types of Span-Oriented Information Extraction Tasks}

There are a variety of information extraction tasks that are defined based upon the needs of the system and the data available. These tasks have been given various names and descriptions in the literature, but they all share the same basic definition of a span.

Given a sentence in some document, for example the sentence, "Apple CEO Tim Cook sold his Texas house", different varieties of information extraction would seek to label different sub-sequences having one or more classes, \eg, Apple as a company, Texas a US state.
Among these varieties are: (1) Entity disambiguation (ED)~\cite{Hoffart-EMNLP'11-robust}, (2) Entity Linking (EL)~\cite{Kolitsas-CoNLL'18-end2end}, (3) Entity Typing (ET)~\cite{Huang-EMNLP'21-few_shot_ner}, (4) Named Entity Recognition (NER)~\cite{Sang-NAACL'03-conll03}, (5) Attribute Value Extraction (AVE)~\cite{Zheng-KDD'18-opentag}, (6) Machine Reading Comprehension (MRC)~\cite{Rajpurkar-EMNLP'16-squad}.

\setlength{\tabcolsep}{2pt}
\begin{table}[t]
\centering
\footnotesize
\caption{Overview of Span-Oriented Information Extraction Tasks}
{\renewcommand{\arraystretch}{1.1}
\input{table/task}
}
\label{tab:task}
\end{table}

\setlength{\tabcolsep}{6pt}

Table~\ref{tab:task} provides a non-exhaustive list of different information extraction tasks and a specific example of that task.


\subsection{Entity Disambiguation} In cases where the surface form is given, either through text matching or some other entity identification task, what remains is to match the surface form with the appropriate class label. In the following examples we identify the surface form with a beginning and end token-index of the input where the first token is indexed at 0.

\begin{table}[h!]
    \centering
    \footnotesize
    \begin{tabular}{>{\raggedleft\arraybackslash}p{2cm} p{3.5in}}
    TASK:&Entity Disambiguation\\
    INPUT:&Apple CEO Tim Cook sold his Texas house\\
    INPUT:&[$(2, 3, ?)$, $(6, 6, ?)$]\\
    OUTPUT:&[$(0, 0, $~\url{wiki/Apple_Inc.}$)$, $(2, 3, $~\url{wiki/Tim_Cook}$)$, $(6, 6, $~\url{wiki/Texas}$)$] \\
    \end{tabular}
\end{table}
Entity disambiguation is so named because the task is mostly to determine which specific entity, if there are many similarly-named entities, that surface form represents. To do this, ED systems rely on the context to make their decisions. The difficulty of ED is that entities typically have a giant number of classes. Subsequently, rare and infrequent entities are difficult to disambiguate~\cite{Wu-EMNLP'20-scalable, Ayoola-NAACL'22-refined, Shen-TKDE'21-survey, sevgili-SW'22-survey}.

\subsection{Entity Linking} In cases where the surface form is not pre-defined, the position of the spans and the class of the span (\ie, the entity) must both be extracted. Compared to ED, EL is much more difficult. EL requires jointly identifying non-standard surface forms from the input text and assigning the correct labels. As a result ED systems are typically more precise, while EL systems observe redundant performance drops in the same datasets~\cite{DeCao-ICLR'21-GENRE, Kolitsas-CoNLL'18-end2end, sevgili-SW'22-survey,Hulst-SIGIR'20-rel}.

\begin{table}[h!]
    \centering
    \footnotesize
    \begin{tabular}{>{\raggedleft\arraybackslash}p{2cm} p{3.5in}}
    TASK:&Entity Linking\\
    INPUT:&Apple CEO Tim Cook sold his Texas house\\
    OUTPUT:&[$(0, 0, $\url{wiki/Apple_Inc.}$)$, $(1, 1, $\url{wiki/Chief_executive_officer}$)$, \newline $(2, 3, $~\url{wiki/Tim_Cook}$)$, $(6, 6, $~\url{wiki/Texas}$)$] \\
    \end{tabular}
\end{table}

The EL task essentially performs the surface form identification sub-task and the ED task simultaneously. This provides greater freedom to the system  so that additional context of any found-span might be used to find more spans. As a result, EL systems generally have greater coverage, but at the expense of precision.

\subsection{Entity Typing} There are also cases where the end-user does not seek a specific entity-entry in some knowledge base, but rather seeks to know the specific (\ie, fine-grained) types of entities that are resident in some span~\cite{Choi-ACL'18-entity_type, Onoe-AAAI'20-fine}.

\begin{table}[h!]
    \centering
    \footnotesize
    \begin{tabular}{>{\raggedleft\arraybackslash}p{2cm} p{3.5in}}
    TASK:&Entity Typing\\
    INPUT:&Apple CEO Tim Cook sold his Texas house\\
    INPUT:&[$(0, 0, ?)$, $(2, 3, ?)$, $(6, 6, ?)$]\\
    OUTPUT:&[$(0, 0, $~\textsf{Company}$)$, $(2, 3, $~\textsf{Businessman}$)$, $(6, 6, $~\textsf{State}$)$] \\
    \end{tabular}
\end{table}

The ET task is a slightly relaxed form of the ED task, where the number of classes is not as large, but can still be extensive depending on the type-granularity that the user seeks. Like in the ED task, because the beginning and ending indexes of the spans are provided as input, this task typically has high precision at the expense of coverage. However, evaluation of the ET task can be difficult because properly aligning the right span-label that matches the ground-truth granularity can be difficult. Indeed, previous work has found that reasonable (and sometimes arguably more-correct span-labels) are often counted as incorrect in ET evaluation~\cite{Choi-ACL'18-entity_type}.

\subsection{Named Entity Recognition} One of the first information extraction tasks from the MUC workshops described above was the NER task. This task seeks to identify entities from a sentence and, almost as a bonus, also labels the entities into one of three or four broad types; typically \texttt{PER}, \texttt{ORG}, and \texttt{LOC} representing person-, organization-, location-entities respectively.

\begin{table}[h!]
    \centering
    \footnotesize
    \begin{tabular}{>{\raggedleft\arraybackslash}p{2cm} p{3.5in}}
    TASK:&Named Entity Recognition\\
    INPUT:&Apple CEO Tim Cook sold his Texas house\\
    OUTPUT:&[$(0, 0, $~\texttt{ORG}$)$, $(2, 3, $~\texttt{PER}$)$, $(6, 6, $~\texttt{LOC}$)$] \\
    \end{tabular}
\end{table}

The primary difficulty of the NER task is the identification of the span's starting and ending indexes; once the surface form can be identified, the class label is rather straightforward because the class label set is typically very small and contains only coarse-grained entity types. However, because NER systems are typically trained on a small set of coarse-grained entity types, many spans that would be easily discovered with EL systems (\eg, animals, technology, works of art) are not easily binned into one of the coarse types. 

\subsection{Attribute Value Extraction} In instances where many sentences describe a general set of items, like, for example, descriptions of episodes on a streaming service, item descriptions in an online game, or groceries available on an e-commerce website, the entities themselves are not of interest. Instead, the AVE task seeks to extract the collective values corresponding to attributes of interest from the descriptive document(s). Nevertheless, AVE can still be views as a span-oriented information extraction task.  Continuing the running example, if we imagine a collection of sentences discussing technology news, then the AVE task might extract the corresponding technology leaders as follows:

\begin{table}[h!]
    \centering
    \footnotesize
    \begin{tabular}{>{\raggedleft\arraybackslash}p{2cm} p{3.5in}}
    TASK:&Attribute Value Extraction\\
    INPUT:&Satya Nadella says that Microsoft products will soon connect to OpenAI. \newline Apple CEO Tim Cook sold his Texas house. \newline Jensen Huang, head of NVIDIA, announces the launch of DGX GH200.
    \\
    OUTPUT:&[$(0, 1, $~\textsf{CEO}$)$, $(13, 14, $~\textsf{CEO}$), (19, 20, $~\textsf{CEO}$)$] \\
    \end{tabular}
\end{table}




\subsection{Machine Reading Comprehension} In some cases users seek to extract spans related to some free-text question. Although the MRC task is unlike many span-oriented information extraction tasks, it still requests the same fundamental output: a span of tokens and a class.

\begin{table}[h!]
    \centering
    \footnotesize
    \begin{tabular}{>{\raggedleft\arraybackslash}p{2cm} p{3.5in}}
    TASK:&Machine Reading Comprehension\\
    INPUT:&Apple CEO Tim Cook sold his Texas house.\\
    INPUT:&Q: Who is the CEO of Apple?\\
    OUTPUT:&[$(2, 3, $~\texttt{Who is the CEO of Apple?}$)$] \\
    \end{tabular}
\end{table}

Like in the AVE task, the MRC task contains a beginning- and ending-index and as well as a class. In this particular case, the class is a direct restating of the input question. This is an important consideration. This may be best explained with an analogy to the NER task. If we re-consider the NER task to be an MRC task, then the question asked of the NER system is the entity-label:

\begin{table}[h!]
    \centering
    \footnotesize
    \begin{tabular}{>{\raggedleft\arraybackslash}p{2cm} p{3.5in}}
    TASK:&Named Entity Recognition\\
    INPUT:&Apple CEO Tim Cook sold his Texas house\\
    INPUT:&Q: PER\\
    OUTPUT:&[$(2, 3, $~\texttt{PER}$)$] \\
    \end{tabular}
\end{table}

Here we see that the answer to the MRC task is not the class label, but is instead denoted by the span indices which reveal the answer to be \texttt{Tim Cook}, who is in Person. 

The above list of span-oriented information extraction tasks is by no means exhaustive, but these examples are meant to be representative of our philosophy: by re-imagining information extraction tasks as systems that output spans, then these systems can be considered natural analogs of one another. With this in mind, the means by which these various information extraction systems are evaluated can be viewed from a more coherent perspective too.


\section{Evaluation of Span-Oriented Information Extraction}
The most common way to evaluate information extraction systems is to use the standard precision, recall and $F_1$ metrics~\cite{Shen-TKDE'14-survey}. However, any metric that evaluates spans deserve a more-thorough consideration. This is because the groundtruth span may not exactly line up with the predicted span, yet still be close-enough to warrant a true-positive judgement. Likewise, the class-label(s) within the predicted spans might not exactly match the groundtruth span, yet still be close-enough to warrant leniency. Because of these consideration several metrics and metric-variants have been developed to handle these difficult cases.


The $F_\beta$ metric was originally intended as a way to balance the precision and recall (\ie, coverage) of machine learning system. When the $\beta=1$ then the precision and recall values are evenly weighted in the $F$-score. Lower values for $\beta$ give more weight to the precision metric and \textit{vice versa}. Unless otherwise specified, systems typically set $\beta=1$ yielding the well-known $F_1$-score. 

The $F_1$-score uses precision and recall metrics, which themselves require some notion of binary true and false predictions. As applied to span-oriented information extraction, this creates a rigid requirement that any token can belong to at most one span and that a span must exactly match the groundtruth to be considered a true-positive instance.

In the simple case where there are only labels (\eg, yes/no, on/off), the $F_1$-score provides a meaningful, although rigid evaluation metric. However, as noted above, most of the class-sets in information extraction are enormous, having thousands (or hundreds of thousands) of classes. In such cases, a decision needs to be made on how to calculate certain mean-averages. The two most common decisions are called (1) the micro $F_1$-score and (2) the macro $F_1$-score.

\subsection{Micro $F_1$-Score}

The micro $F_1$ score is widely recognized as the standard evaluation metric for rigid NER. The notion of rigidity in this instance denotes that any token of the input document can belong to at most one span and that a true-positive instance must match the groundtruth span exactly.

Formally, for each class $c^*\in C$ and for a groundtruth set $s=(b,e,c)\in S$ and for a set of predicted span instances $\hat{s}=(\hat{b}, \hat{e}, \hat{c})\in \hat{S}$, the number of true positives ($\textrm{TP}_{c^*}$) is:

\begin{equation} \label{equation:true_positive_class}
    \textrm{TP}_{c^*} = \sum_{(b,e,c)\in S} \sum_{(\hat{b}, \hat{e}, \hat{c})\in \hat{S}} \mathbbm{1} \left( b=\hat{b} \land e=\hat{e} \land c = \hat{c} = c^* \right)
\end{equation}




Likewise we count the number of false negatives ($\textrm{FN}_{c^*}$) and false positives $\textrm{FP}_{c^*}$ as:

\begin{equation}
\label{equation:false_negative_class}
    \textrm{FN}_{c^*} = \sum_{(b,e,c)\in S} \left( 1 - \sum_{(\hat{b}, \hat{e}, \hat{c})\in \hat{S}} \mathbbm{1} \left(b=\hat{b} \land e = \hat{e} \land c = \hat{c} = c^* \right) \right)
\end{equation}

\begin{equation}
\label{equation:false_positive_class}
    \textrm{FP}_{c^*} = \sum_{(\hat{b},\hat{e},\hat{c})\in \hat{S}} \left( 1 - \sum_{({b}, {e}, {c})\in {S}} \mathbbm{1} \left(b=\hat{b} \land e = \hat{e} \land c = \hat{c} = c^* \right) \right)
\end{equation}

Note that some tasks, like MRC and AVE sometimes relax the indicator function $\mathbbm{1}(\cdot)$ so that the beginning- and ending-indexes need not exactly match, but that the tokens denoted by these indexes match: $\mathbbm{1} (\textrm{substr}(b,e) = \textrm{substr}(\hat{b},\hat{e}) \land c = \hat{c} = c^*)$. For example, in MRC we do not need to know that \texttt{Tim Cook} begins and ends with tokens 2 and 3, only that the tokens between indexes 2 and 3 match the groundtruth answer.

Then, to obtain the micro $F_1$-scores the TPs, FNs, and FPs are summed across the various classes and substituted into the standard precision and recall metrics to obtain micro-precision and micro-recall. 

\begin{equation}
    \label{equation:micro_metric}
    \textrm{micro-Prec} = \frac{ \sum_{c^*\in C} \textrm{TP}_{c^*} }{  \sum_{c^*\in C} (\textrm{TP}_{c^*} + \textrm{FP}_{c^*}) }; \textrm{micro-Rec} = \frac{ \sum_{c^*\in C} \textrm{TP}_{c^*} }{  \sum_{c^*\in C} (\textrm{TP}_{c^*} + \textrm{FN}_{c^*}) }
\end{equation}

\begin{equation}
\textrm{micro-}F_1 = \frac{2\times \textrm{micro-Prec}\times \textrm{micro-Rec}}{\textrm{micro-Prec} + \textrm{micro-Rec}}
\end{equation}


\subsection{Macro $F_1$-Score}
The micro $F_1$-score is a natural way to sum up the successes and errors of the model predictions. However, this simple solution can be easily swayed in the likely case that the class labels are unbalanced, \ie, certain labels occur much more frequently than others. In this situation, a class-based precision and recall measurement can be calculated as follows:

\begin{equation}
\label{equation:macro_metric}
    \textrm{Prec}_{c^*} = \frac{\textrm{TP}_{c^*}}{\textrm{TP}_{c^*} + \textrm{FP}_{c^*}}; \textrm{Rec}_{c^*} = \frac{\textrm{TP}_{c^*}}{\textrm{TP}_{c^*} + \textrm{FN}_{c^*}}; F_{1,c^*} = \frac{2 \times \textrm{Prec}_{c^*} \times \textrm{Rec}_{c^*}}{\textrm{Prec}_{c^*} + \textrm{Rec}_{c^*}}
\end{equation}

Then, the overall macro $F_1$-score is the arithmetic mean of $F_1$ scores across all the individual classes as follows:

\begin{equation}
\label{equation:macro_f1}
    \textrm{macro-}F_{1} = \frac{\sum_{c^*\in C}F_{1,c^*}}{|C|}
\end{equation}

\subsection{Exact Match Evaluation}
The macro and micro metrics described above require exact matches of the beginning-index, ending-index, and the class $\mathbbm{1} (b=\hat{b} \land e = \hat{e} \land c = \hat{c} = c^* )$ in order to count towards a true positive instance. 

Also note that the string matching function $\mathbbm{1} (\textrm{substr}(b,e) = \textrm{substr}(\hat{b},\hat{e}) \land c = \hat{c} = c^*)$ commonly used in the AVE and MRC tasks do not require exact matches of the indices, but do require exact matches of the sub-sequences represented by the indexes.

\begin{table}[h!]
    \centering
    \footnotesize
    \begin{tabular}{>{\raggedleft\arraybackslash}p{2cm} p{3.5in}}
    TASK:&NER/MRC\\
    INPUT:&Apple CEO Tim Cook sold his Texas house\newline Tim Cook announces new M2 chip.\\    
    GT:&[$(2, 3, $~\texttt{PER}$)$] \\
    OUTPUT:&[$(8, 9, $~\texttt{PER}$)$] \\
    \end{tabular}
\end{table}

In the example directly above the extracted span $(8, 9, $~\texttt{PER}$)$ does match ground truth span in the MRC and AVE task, but would not match the ground truth (GT) span in NER, EL, ET, and ED tasks.

\subsection{Relaxed Match Evaluation}
Exact matching requirements are often criticised for imposing too strict of a requirement onto the system. It is often the case that a sub-sequence or super-sequence of the ground truth span is an equally valid match. Likewise, in fine-grained ET or ED tasks, a close, but still inexact match between the predicted class $\hat{c}$ and the ground truth class $c^*$ could also be equally valid (and our experience shows that sometimes the predicted match is arguably better than the ground truth match)~\cite{Ding-NAACL_workshop'22-posthoc}. To allay this criticism, the use of relaxed (\ie, partial) span matching is also used in evaluation~\cite{Kolitsas-CoNLL'18-end2end, Shen-TKDE'21-survey, Grishman-COLING'96-MUC-6, Doddington-LREC'04-ACE}. Under a relaxed span matching regime, if a predicted span has the same class as one of the ground truth spans and the span indices intersect, then that prediction is counted as a true positive.

\begin{table}[h!]
    \centering
    \footnotesize
    \begin{tabular}{>{\raggedleft\arraybackslash}p{2cm} p{3.5in}}
    TASK:&NER\\
    INPUT:&Apple CEO Tim Cook sold his Texas house\newline Tim Cook announces new M2 chip.\\    
    GT:&[$(2, 3, $~\texttt{PER}$)$] \\
    OUTPUT:&[$(3, 6, $~\texttt{PER}$)$] \\
    \end{tabular}
\end{table}

In the example directly above, the extracted span representing the subsequence \texttt{Cook sold his Texas} would count as a true positive for the ground truth span representing the sequence \texttt{Tim Cook}. This additional tolerance almost always results in a positive performance shift; however, partial matching may sometimes be too lenient---as shown in the example above. Therefore, strict span matching remains the standard regime used to evaluate span-oriented information extraction tasks.



\section{Taxonomy of Information Extraction Features}

In this section, we begin to consider how information extraction systems use natural language to create a model from which information can be extracted. To that end, we will summarize the features commonly gleaned from (1) tokens, (2)  spans, and (3) span classes. 

Because natural language is digitally represented as a sequence of bytes in its most basic form, we consider that to be the lowest-level representation of written (digital) language. From that form natural language tokenizers turn bytes into words (or sub-words) from which sentences, Tweets, paragraphs, posts, articles, stories, and narratives are formed. Information extraction tasks typically operate at the token-level; by our definition, these systems output a span, which is a sequence of one or more words and a class. Each of these levels: the token, the span, and the class all have information that can be used in the construction of a natural language model. This section will briefly highlight each.

\subsection{Token Features}
Most natural language is grouped into tokens---typically words. These tokens are fundamental element in communication; dictionaries, for example, are one source of understanding for these tokens, as are encyclopedias and thesauri. The same is true in natural language processing. Because tokens are the basic elements, crafting the corresponding token features to be flexible and generalized is an important consideration for most tasks.
\subsubsection{Linguistic Token Features}
Many token features are linguistic in their nature. For example, part-of-speech tags are one of the earliest token features used to distinguish word classes (\eg, noun, verb, adjective, and adverb)~\cite{Kroeger-BOOK'05-analyzing}. The abstract syntax tree is another linguistic feature that transforms plain text into a self-referential tree structure~\cite{Neamtiu-IWMSR'05-understanding}. These approaches produce symbolic properties and are easily interpreted by human beings. However, linguistic features have three significant limitations. First, they do not directly provide information of interest in to most practical applications. For example, although knowing whether a token is a noun or a verb or modifies some other token can be useful in downstream tasks, this is not directly useful in many applications.
Second, training linguistic models requires an enormous amount of expert human annotations~\cite{Chen-EMNLP'14-fast_parsing, Marcus-TreeBanks'93-penn_treebank}. In the decades since linguistic token features were first proposed, many datasets have been created, but these features constantly require updating. Third, even perfect labels result in performance limitations on many IE tasks~\cite{Chen-EMNLP'14-fast_parsing, Yamada-EMNLP'20-wikipedia2vec, Yamada-EMNLP'20-luke}. This is because shallow, token-based information represents a limited view of the deeper intent and meaning within natural language~\cite{Li-TKDE'21'-NER_Survey, Shen-TKDE'14-survey}. 

\subsubsection{Pre-trained Token Features from Language Models}
With the development of language models (LMs), pre-trained word embeddings have become a primary source of token features. The goal of any LM (large or small) is to model the probability distribution over sequences of tokens. That is, given a document $d$ composed of a sequence of tokens $d = \langle t_1, t_2, \ldots, t_c, \ldots, t_{\ell(d)}\rangle$, an LM provides for the estimation of the probability distribution of any token $t_c$ by utilizing other contextual words in the sequence as follows:

\begin{equation}
    p\left( t_c|t_1, t_2, \ldots, t_{c-2}, t_{c-1}, t_{c+1}, t_{c+2}, \ldots, t_{\ell(d)}\right)
\end{equation}

Constructing LMs has been one of the most fundamental and important tasks for the NLP community. So called \textit{large} LMs (LLMs) are able to scale because they are trained in a self-supervised regime without any human annotation at all. As a result, LLMs have shown the ability to learn token features from a wide variety of documents from different domains. Early pre-trained word embeddings were based on the bag-of-word or skip-gram models; which is best represented by word2vec~\cite{Mikolov-NIPS'13-word2vec}, GloVe~\cite{Pennington-EMNLP'14-glove} and fasttext~\cite{Bojanowski-TACL'17-fasttext}. With the development of text Transformers, self-supervised token features has been widely adopted and even supplanted most alternatives in natural language processing. The two most representative projects in this category are (1) the GPT series having a left-to-right language model~\cite{Radford-Technique_Report'18-gpt,Radford-Technique_Report'19-gpt2, Brown-NIPS'20-gpt3} and (2) the BERT series with masked language modeling (MLM)~\cite{Devlin-NAACL'19-bert}. The broad pre-training with self-supervised labels that these LLMs undergo is typically sufficient for most tokens to obtain robust features~\cite{Mikolov-NIPS'13-word2vec, Devlin-NAACL'19-bert}. Therefore these pre-trained LLMs can also be adapted to tackle information extraction tasks. This is also important because information extraction tasks typically have limited and biased training data~\cite{Sang-NAACL'03-conll03, Doddington-LREC'04-ACE}, and it is difficult and even inapplicable to employ self-supervised training to the information extraction tasks~\cite{Devlin-NAACL'19-bert, Brown-NIPS'20-gpt3, Lu-ACL'22-UIE}. 


\subsubsection{Character Features}
Our definition of a span uses tokens as the base type. However, in some cases the tokens are unable to provide a useful properties of characters. In these cases, it may be beneficial to consider individual characters as extra supplemental features. These character features can then be used as extra learning parameters to improve performance in some specific instances including Chinese language modeling~\cite{Meng-NIPS'19-glyce, Sun-ACL'21-chinesebert}, and in fields that commonly use acronyms and initialisms like chemistry~\cite{Ding-BigData'19-chemistry, Wang-EMNLP'21-chemner}, biology~\cite{Lee-Bioinformatics'20-biobert}, and law~\cite{Chalkidis-EMNLP'20-legal}.

\subsubsection{Token Sequences}
Because natural language is (digitally) expressed in a sequence of bytes or tokens, there has been a large effort to model these sequences~\cite{Li-TKDE'21'-NER_Survey, sevgili-SW'22-survey}. Arguably the most well known model for token sequences is the transformers~\cite{Vaswani-NIPS'17-attention, Devlin-NAACL'19-bert}, although other architectures like the recurrent neural network (RNN)~\cite{hochreiter1997long, Ma-EMNLP'17-ner_bi_rnn}, convolutional neural network (CNN)~\cite{lecun1995convolutional,McCallum-EMNLP'17-iterated_dilated_cnn}, and the point network~\cite{Zhou-AAAI'17-point_ner, Vinyals-NIPS'15-point_networks} have been used as well. Alternatives to these neural models tend to use probability graphic models to model token dependencies with implementations such as the hidden Markov model (HMM)~\cite{Morwal-IJNLC'12-hmm}, the maximum entropy Markov model (MeMMs)~\cite{McCallum-ICML'00-memm} and the conditional random field (CRF)~\cite{Lafferty-ICML'01-crf}. 

\subsection{Span Features}
\subsubsection{Span Embeddings}
A span defined as sequence of one or more tokens may therefore have a variety of representations. Typically, span embeddings are  built on top of token features. For example, Chen et al. directly used the word embedding of the first token to represent the span features~\cite{Chen-AACL'20-contextualized}, and Tan et al. used the concatenation of word embeddings of the first and last tokens~\cite{Tan_AAAI'20-boundary}. Otherwise, single-pass frameworks use an averaging over a pool of token embeddings to form span embeddings~\cite{Ayoola-NAACL'22-improving, Ayoola-NAACL'22-refined}.
The PURE model further learns a length embedding as part of span embeddings~\cite{Zhong-NAACL'21-pure}. The W2NER model uses a convolutional layer and an LSTM jointly to fuse token embeddings into span embeddings~\cite{Li-AAAI'22-w2ner}. 

In addition to formulating span embeddings based on token embeddings, another approach is to create span embeddings from scratch as a different pre-training task. Early work on this trajectory extends the word2vec idea to learn span embeddings from contextual span-token and span-span contextual correlations~\cite{Yamada-CoNLL'16-joint, Yamada-EMNLP'20-wikipedia2vec}. Later work in this area extended the mask language model of BERT to generate span embeddings as a co-training task within language modeling~\cite{Broscheit-CoNLL'19-investigating, Yamada-EMNLP'20-luke, Chen-AAAI'20-improve, Yamada-NAACL'22-global}.

\subsubsection{Span Sequences}
Just as token features can be modelled as token sequences, span embeddings can likewise be modelling as a sequence of spans. Understanding span sequences is critical in many information extraction tasks. For example, in entity disambiguation one of the primary features used to select span candidates is the context-tokens and other neighbor spans that surround the span in the same sentence. For example, the tokens \texttt{Apple} and \texttt{CEO} in the running example in Section 2 could help to disambiguate \texttt{Tim Cook} as \url{wiki/Tim\_Cook} from some another entry with a similar name. Likewise, the identification of \url{wiki/Tim_Cook} may aid in the recognition that \texttt{Apple} refers to \url{wiki/Apple\_Inc.} and not some other entity. Formally, most previous frameworks consider the training objective of a common classification model~\cite{ganea-EMNLP'17-deep_ed, Kolitsas-CoNLL'18-end2end, Phan-CIKM'17-neupl, Phan-TKDE'18-pair} adapted in Eq.(\ref{entity_linking_global}):

\begin{equation}
g(\textbf{s, c}) = \sum_{i=1}^{n}\Phi(s_i, c_i) + \sum_{i < j} \Psi(c_i, c_j) \label{entity_linking_global}
\end{equation}

\noindent where the contextual span scores $\Phi(s_i, c_i)$ for each span $s_i$ and class $s_i$ and the span-span correlation scores $\Psi(c_i, c_j)$ for the predicted classes $c_i$ and $c_j$ are both used to train the model. Modelling span sequences tends to work well when the number of classes is small. In this scenario, adjacent and correlated tokens are typically sufficient to determine the span class. However, when the number of classes is large like in the fine-grained ET task and the EL task, contextual words may not be enough to determine the corresponding span classes\cite{ganea-EMNLP'17-deep_ed, Kolitsas-CoNLL'18-end2end, raiman-AAAI'18-deeptype, raiman-AAAI'22-deeptype2}.

\subsection{Span Class Representations}
Most information extraction models produce spans where the class element is a simple id or perhaps a pointer (\eg, \texttt{PER}, \texttt{businessman}, \url{wiki/Texas}). However, many span classes contain metadata such as text descriptions and even relationship information (\eg, in the case of Wikipedia). These span classes can be used to add additional context to possibly improve performance. Yet another option is to use pre-trained methods to obtain representations for each span's class. For example, the TagMe system used hyperlinks among pages to learn a class representation~\cite{Ferragina-CIKM'10-tagme}; likewise, wikipedia2vec~\cite{Yamada-EMNLP'20-wikipedia2vec} and deep-ed~\cite{ganea-EMNLP'17-deep_ed} extended word2vec to learn token and class correlations. In a similar way, ERNIE~\cite{Liu-ACL'19-ERNIE} and LUKE~\cite{Yamada-EMNLP'20-luke} extended the BERT masking language model to entity disambiguation in order to obtain better token and span class representations. 


\section{Model Transformations among Information Extractors}


Having previously identified the various information extraction tasks and their features, the next piece of the puzzle is to describe how spans are transformed by different information extraction models. This section presents a different, yet unifying, perspective on information extraction by considering the various \textit{transformations} that a span undergoes for different information extraction tasks.

\newcommand{\midsepremove}{\aboverulesep = 0mm \belowrulesep = 0mm}
\newcommand{\midsepdefault}{\aboverulesep = 0.605mm \belowrulesep = 0.984mm}
\newcolumntype{P}[1]{>{\centering\arraybackslash}p{#1}}
\newcommand{\bc}[1]{\multicolumn{1}{c|}{\cellcolor{blue!20}#1}}
\newcommand{\pc}[1]{\multicolumn{1}{c|}{\cellcolor{purple!20}#1}}
\newcommand{\ec}[1]{\multicolumn{1}{c|}{#1}}
\newcommand{\lc}[1]{\multicolumn{1}{c}{#1}}
\newcommand{\dl}[1]{\multicolumn{1}{P{1.4cm}!{\color{gray}\vrule width 0.01em}}{#1}}

In their attempt to tackle different information extraction tasks, different information extraction models employ various \textit{transformations} to the spans. These transformations can be grouped into six types of transformations: (\ref{sec:seqlab}) sequential labeling, (\ref{sec:tokpot}) token prototyping, (\ref{sec:tokpair}) token-pair transformation, (\ref{sec:spanclass}) span classification, (\ref{sec:spanloc}) span locating, and (\ref{sec:spangen}) span generation.

\begin{table}[t]
\centering
\footnotesize
\caption{Matrix of Information Extraction Tasks by their Transformation Type}
\begin{tabular}{@{}r| P{1.5cm}P{1.4cm}P{1.5cm}P{1.3cm}P{1.3cm}P{1.5cm}@{}}
\toprule
Transformation & NER & ED & EL & ET & AVE & MRC \\
\midrule
Sequential Labeling  &\dl{\cite{Huang-15-bidirectional, Ma-ACL'16-end, Devlin-NAACL'19-bert, Yamada-EMNLP'20-luke}} & \dl{~} &\dl{\cite{Hulst-SIGIR'20-rel, Chen-AACL'20-contextualized, Ayoola-NAACL'22-refined, Kolitsas-CoNLL'18-end2end}} & \dl{~} & \dl{\cite{Yan-ACL'21-adatag,Zheng-KDD'18-opentag,Xu-ACL'19-suopentag}} & \cite{Dhingra-ACL'17-gated_mrc,Sordoni-arxiv'16-iterative_mrc, Kadlec-ACL'16-mrc}\\[.2cm] \arrayrulecolor{gray}\cline{2-7}
Token Prototype  & \dl{\cite{Huang-EMNLP'21-few_shot_ner}} & \dl{\cite{Broscheit-CoNLL'19-investigating, Yamada-NAACL'22-global}}  & \dl{\cite{Broscheit-CoNLL'19-investigating, Yamada-NAACL'22-global,ganea-EMNLP'17-deep_ed, Yamada-EMNLP'20-wikipedia2vec, Yamada-EMNLP'20-luke}} &\dl{ \cite{Ma-COLING'16-label_embedding_et}} & \dl{\cite{Yan-ACL'21-adatag, Zalmout-EMNLP'22-prototype}} & \\\cline{2-7}
Token-pair Classification  &\dl{\cite{Li-AAAI'22-w2ner}} & \dl{}&\dl{} &\dl{} &\dl{} &  \\[.3cm]\cline{2-7}
Span Classification &\dl{\cite{Zeng-EMNLP'20-tri, Angell-NAACL'21-clustering, NegSampling-ICLR'21-Shi, Tan_AAAI'20-boundary, Sohrab-EMNLP'18-exhaustive, Zhong-NAACL'21-pure}}  & \dl{\cite{Wu-EMNLP'20-scalable, Agarwal-NAACL'22-arboel, Yamada-EMNLP'20-wikipedia2vec, ganea-EMNLP'17-deep_ed, raiman-AAAI'18-deeptype, raiman-AAAI'22-deeptype2, Chen-AACL'20-contextualized}}  &  \dl{\cite{Kolitsas-CoNLL'18-end2end, Hulst-SIGIR'20-rel, Chen-AAAI'20-improve}} &\dl{\cite{Choi-ACL'18-entity_type, Corro-EMNLP'15-FINET, Murty-ACL'18-hierarchical_et_el}} & \dl{\cite{Ding-EMNLP'22-ask}} &\cite{Back-ICLR'20-neurquri,Min-ACL'19-multi_hop_rc,Yu-arxiv'16-DCR}\\\cline{2-7}
Span Locating  &\dl{\cite{Li-ACL'20-unified, Shen-ACL'21-locate_and_label} } &\dl{\cite{Barba-ACL'22-extend, Gu-AAAI'21-m3} } & \dl{\cite{zhang-ICLR'22-entqa}} & \dl{~} & \dl{\cite{Wang-KDD'20-aveqa, Ding-EMNLP'22-ask}}  &\cite{Zhang-AAAI'21-retrospective_mrc, Seo-ICLR'17-mrc} \\[.3cm]\cline{2-7}
Span Generation  &\dl{ \cite{Lu-ACL'22-UIE, Fei-NIPS'22-LasUIE, Yan-ACL'21-bertner, Shang-EMNLP'22-formulate}} & \dl{\cite{DeCao-ICLR'21-GENRE}} & \dl{\cite{DeCao-ICLR'21-GENRE}} &\dl{\cite{Ding-arxiv'21-prompt_et}} & \dl{} &\cite{Qi-EMNLP'20-prophetnet, Wang-ICLR'17-mrc, Lewis-ACL'20-bart} \\

\bottomrule
\end{tabular}
\label{tab:transforms}
\end{table}

As we shall see, unifying these different tasks reveals the importance of the various transformations. For example, the sequential labelling transformation commonly used in the NER task  appears to be vastly different than the two-step transformation used in the EL task. However, as we shall see, because these tasks share the same input and output, these transformations do naturally generalize to other transformations and are actually swappable. Despite their interchangeability, the taxonomy of different transformations (see Tab.~\ref{tab:transforms} for details) does come with trade offs for different tasks. For example, the sequential labelling transformation decomposes span-labels into token-labels where each token assigned a label. This transformation ignores token locality features, which could be important in accurately finding span boundaries, but is nevertheless fast and easy to train. In this section we describe different model transformations and briefly discuss their trade offs.



\subsection{Transformation Approaches}

\subsubsection{Sequential labeling} 
\label{sec:seqlab}

Sequential labeling (\ie, token classification) is the most traditional and common transformation used in NER~\cite{Huang-15-bidirectional} as well as other information extraction tasks~\cite{Zheng-KDD'18-opentag, Yan-ACL'21-adatag, Ayoola-NAACL'22-refined}. 
The core idea of sequential labelling is to directly transform spans into token-wise classes labeled with the Begin, Inside, Outside, End (BIOE) schema, where each label represents a token that begins, is inside of, ends, or is outside of some span. This schema has been expanded to also include other labels, like Left and Right (L/R), to represent tokens to the left and right of a span~\cite{Li-TKDE'21'-NER_Survey}.

\begin{table}[h!]
    \centering
    \footnotesize
    \begin{tabular}{>{\raggedleft\arraybackslash}p{2.5cm} cccccccc}
    TASK:&\multicolumn{5}{l}{NER + Sequential Labelling}\\[.1cm]
    INDEX:&0 & 1& 2 &3 &4 &5 &6& 7\\    
    INPUT:&Apple & CEO& Tim &Cook &sold &his &Texas& house\\    
    TRANSFORM:&B-ORG & O& B-PER & E-PER &O &O &B-LOC & O\\[.1cm]
    OUTPUT:&\multicolumn{5}{l}{[$(0, 0, $~\texttt{ORG}$)$, $(2, 3, $~\texttt{PER}$)$, $(6, 6, $~\texttt{LOC}$)$]} \\
    \end{tabular}
\end{table}

Continuing the example above, the sequential labelling transformation uses the B and E labels to identify the beginning and end of the span encompassing \texttt{Tim Cook} one token at a time. Those tokens that are outside of the span are labeled with O. Then, during inference, token classes are first labeled and then span labels are obtained by concatenating one or more continuous tokens belonging to the same class. Note that the I and E labels can be missing in a span when a span has only one or two tokens.



\subsubsection{Token Prototyping}
\label{sec:tokpot}
Token prototyping considers each span to be a sequence of tokens, which is then mapped to the same class~\cite{Yan-ACL'21-adatag, Zalmout-EMNLP'22-prototype, Huang-EMNLP'21-few_shot_ner}. Unlike sequential labelling, which labels tokens one at a time, token prototypes (\eg, \texttt{PER}, Businessman) are computed with locality and clustering based objectives. During inference, the tokens classes are represented as prototypes and embeddings are computed for each token. Then tokens that are close in the embedding space are clustered together and the corresponding classes are obtained by selecting the closest prototype.

\begin{table}[h!]
    \centering
    \footnotesize
    \begin{tabular}{>{\raggedleft\arraybackslash}p{2.5cm} cccccccc}
    TASK:&\multicolumn{5}{l}{NER + Token Prototyping}\\[.1cm]
    INDEX:&0 & 1& 2 &3 &4 &5 &6& 7\\    
    INPUT:&Apple & CEO& Tim &Cook &sold &his &Texas& house\\    
    TRANSFORM:&\texttt{ORG} & O& \texttt{PER} & \texttt{PER} &O &O &\texttt{LOC} & O\\[.1cm]
    OUTPUT:&\multicolumn{5}{l}{[$(0, 0, $~\texttt{ORG}$)$, $(2, 3, $~\texttt{PER}$)$, $(6, 6, $~\texttt{LOC}$)$]} \\
    \end{tabular}
\end{table}

Because this is a token-oriented approached, the tokens \texttt{Tim} and \texttt{Cook}, from the example above, are both individually assigned the \texttt{PER} label. Of course, this provides some ambiguity: it is unclear whether \texttt{Tim Cook} is a single person or two persons \texttt{Tim} and \texttt{Cook}. Typically, a post-processing step combines labels of the same type into a single multi-token span, but this isn't always desirable. 

\subsubsection{Token-pair Classification}
\label{sec:tokpair}
In token-pair classification, and as the name suggests, span labels are transformed into relationships between two tokens. 
During inference, each token-pair is labeled as one of an assortment of classes that describes the relationship between the two words within the same span.

\midsepremove
\begin{table}[h!]
\vspace{0.5cm}
    \centering
    \footnotesize
    \begin{tabular}{r r | P{.6cm}P{.6cm}P{.6cm}P{.6cm}P{.6cm}P{.6cm}P{.6cm}P{.6cm}}
    TASK:&\multicolumn{5}{l}{NER + Token-Pair Classification}\\[.15cm]
    INDEX & &0 & 1& 2 &3 &4 &5 &6& 7\\    
    & INPUT:&Apple & CEO& Tim &Cook &sold &his &Texas& house\\   \midrule
    0& \rule{0pt}{.4cm}Apple &\bc{NNW} &\bc{} &\bc{} &\bc{} &\bc{} & \ec{} & \ec{}& \ec{} \\[.15cm] \cmidrule[0.1pt]{3-10} 
    1& \rule{0pt}{.4cm}CEO &\pc{} &\bc{} &\bc{} &\bc{} &\bc{} & \bc{} & \ec{}& \ec{} \\[.15cm] \cmidrule[0.1pt]{3-10} 
    2& \rule{0pt}{.4cm}Tim &\pc{} &\pc{} &\bc{} &\bc{NNW} &\bc{} & \bc{} & \bc{}& \ec{} \\[.15cm] \cmidrule[0.1pt]{3-10} 
    3& \rule{0pt}{.4cm}Cook &\pc{} &\pc{} &\pc{THW} &\bc{} &\bc{} & \bc{} & \bc{}& \bc{} \\[.15cm] \cmidrule[0.1pt]{3-10} 
    4& \rule{0pt}{.4cm}sold &\pc{} &\pc{} &\pc{} &\pc{} &\bc{} & \bc{} & \bc{}& \bc{} \\[.15cm] \cmidrule[0.1pt]{3-10} 
    5& \rule{0pt}{.4cm}his &\ec{} &\pc{} &\pc{} &\pc{} &\pc{} & \bc{} & \bc{}& \bc{} \\[.15cm] \cmidrule[0.1pt]{3-10} 
    6& \rule{0pt}{.4cm}Texas &\ec{} &\ec{} &\pc{} &\pc{} &\pc{} & \pc{} & \bc{NNW}& \bc{} \\ [.15cm]\cmidrule[0.1pt]{3-10} 
    7& \rule{0pt}{.4cm}house &\ec{} &\ec{} &\ec{} &\pc{} &\pc{} & \pc{} & \pc{}& \bc{} \\[.15cm]\cmidrule[0.1pt]{3-10} 
    \rule{0pt}{.6cm}OUTPUT:&\multicolumn{5}{l}{[$(0, 0, $~\texttt{ORG}$)$, $(2, 3, $~\texttt{PER}$)$, $(6, 6, $~\texttt{LOC}$)$]} \\
    \end{tabular}
\vspace{0.5cm}
\end{table}
\midsepdefault

For example, the W2NER model uses labels \textit{Next-Neighboring-Word} (NNW) to describe the relationship between pairs of words within a single span~\cite{Li-AAAI'22-w2ner}. Applying this model to the running example produces the transformation above where blue cells represent the window size permitted by the model. Here the token-pair \texttt{Tim}--\texttt{Cook} is labeled with an NNW class describing \texttt{Cook} as the next-neighboring-word of \texttt{Tim}.

Although this transformation produces a sparse matrix with $\ell^2$ possible labels, it does permit non-contiguous dependency references to be labels produced in certain circumstances. Typically these additional labels are labeled as a \textit{Tail-Head-Word} (THW) in the bottom-diagonal. The above illustration shows an example THW that, depending on the task, might be expanded THW-\texttt{PER} to indicate that the span refers to a person.



\subsubsection{Span Classification}
\label{sec:spanclass}
The core idea of the span classification transformation has two parts: (1) span candidate generation and (2) span label assignment. There are many ways to generate span candidates. The simplest is to enumerate all the possible spans up to certain window-lengths (\ie, n-grams)~\cite{Sohrab-EMNLP'18-exhaustive, Zhong-NAACL'21-pure}. The window-length is normally a constant which is typically less than 5 in most tasks. Another way to generate spans is to learn a specific span generation model. For example, many of the token-oriented transformation approaches can generate span candidates with high-probability boundaries~\cite{Zheng-EMNLP'19-boundary, Tan_AAAI'20-boundary}. In another vein, the Ask-and-Verify model uses a machine reading comprehension module to generate span candidates by finding potential boundaries from tokens with high predicted probabilities~\cite{Ding-EMNLP'22-ask}. Span candidates can also be generated utilizing external data and models. For example, most existing entity disambiguation methods utilize rule based methods such as string match and frequency statistics~\cite{ganea-EMNLP'17-deep_ed,Kolitsas-CoNLL'18-end2end,Hulst-SIGIR'20-rel,le-ACL'18-improving, Chen-AAAI'20-improve, Ayoola-NAACL'22-improving, Ayoola-NAACL'22-refined}. Another popular way is to use retrieval models like TF-IDF~\cite{Angell-NAACL'21-clustering}, BM25~\cite{logeswaran-ACL'19-zero}, phrase-mining~\cite{Shang-TKDE'18-auto_phrase, Zeng-EMNLP'20-tri}, or dense retrieval~\cite{Wu-EMNLP'20-scalable} among many others.

After span candidates are generated, the goal of the following span label assignment step is to find a mapping function to select spans from the most promising span candidates and provide a class label. One common method is to employ a span-oriented classifier to distinguish positive span candidates from negative span candidates by screening all the span candidates~\cite{Sohrab-EMNLP'18-exhaustive, Zheng-EMNLP'19-boundary, Tan_AAAI'20-boundary, Ding-EMNLP'22-ask}. One particularly compelling model, Locate-and-Label, which was inspired by two-stage object detection methods in computer vision also considers partially overlapped span candidates as positive samples as long as the intersection over union (IoU) is larger than a certain threshold value~\cite{Shen-ACL'21-locate_and_label}. 

\midsepremove
\begin{table}[h!]
\vspace{0.5cm}
    \centering
    \footnotesize
    \begin{tabular}{r r | P{.6cm}P{.6cm}P{.6cm}P{.6cm}P{.6cm}P{.6cm}P{.6cm}P{.6cm}}
    TASK:&\multicolumn{5}{l}{NER + Span Classification}\\[.15cm]
    INDEX & &0 & 1& 2 &3 &4 &5 &6& 7\\    
    & INPUT:&Apple & CEO& Tim &Cook &sold &his &Texas& house\\   \midrule
    0& \rule{0pt}{.4cm} Apple &\bc{ORG} &\bc{} &\bc{} &\bc{} &\bc{} & \ec{} & \ec{}& \ec{} \\[.15cm] \cmidrule[0.1pt]{3-10} 
    1& \rule{0pt}{.4cm}CEO &\ec{} &\bc{} &\bc{} &\bc{} &\bc{} & \bc{} & \ec{}& \ec{} \\[.15cm] \cmidrule[0.1pt]{4-10} 
    2& \rule{0pt}{.4cm}Tim &\lc{} &\ec{} &\bc{} &\bc{PER} &\bc{} & \bc{} & \bc{}& \ec{} \\[.15cm] \cmidrule[0.1pt]{5-10} 
    3& \rule{0pt}{.4cm}Cook &\lc{} &\lc{} &\ec{} &\bc{} &\bc{} & \bc{} & \bc{}& \bc{} \\[.15cm] \cmidrule[0.1pt]{6-10} 
    4& \rule{0pt}{.4cm}sold &\lc{} &\lc{} &\lc{} &\ec{} &\bc{} & \bc{} & \bc{}& \bc{} \\[.15cm] \cmidrule[0.1pt]{7-10} 
    5& \rule{0pt}{.4cm}his &\lc{} &\lc{} &\lc{} &\lc{} &\ec{} & \bc{} & \bc{}& \bc{} \\[.15cm] \cmidrule[0.1pt]{8-10} 
    6&\rule{0pt}{.4cm} Texas &\lc{} &\lc{} &\lc{} &\lc{} &\lc{} & \ec{} & \bc{LOC}& \bc{} \\[.15cm] \cmidrule[0.1pt]{9-10} 
    7& \rule{0pt}{.4cm}house &\lc{} &\lc{} &\lc{} &\lc{} &\lc{} & \lc{} & \ec{}& \bc{} \\ \cmidrule[0.1pt]{10-10} 
    \rule{0pt}{.6cm}OUTPUT:&\multicolumn{5}{l}{[$(0, 0, $~\texttt{ORG}$)$, $(2, 3, $~\texttt{PER}$)$, $(6, 6, $~\texttt{LOC}$)$]} \\
    \end{tabular}
    \vspace{0.5cm}
\end{table}
\midsepdefault

\setlength\extrarowheight{0cm}

Continuing the running example, we utilize a upper triangular matrix with blue color to represent valid span candidates and the positive span is labeled with corresponding classes.

\subsubsection{Span Locating} 
\label{sec:spanloc}
Another transformation approach is called span locating. The goal of this transformation is to consider an input sentence and relevant classes as a context-query pair and then find the corresponding span boundaries within the original input sentence~\cite{Li-ACL'20-unified}. The pipeline of this transformation is similar to machine reading comprehension (MRC), which is used to find answers in the context to the corresponding questions~\cite{Rajpurkar-EMNLP'16-squad, Rajpurkar-ACL'18-squad2.0}. Unlike in the span classification transformation where span candidates are generated, selected and labeled, in span locating the decision making process happens in reverse order: first the class label is determined and then the proper span holding that label is found. 

\begin{table}[h!]
    \centering
    \footnotesize
    \begin{tabular}{>{\raggedleft\arraybackslash}p{2.5cm} P{.7cm}cccccP{.7cm}c}
    TASK:&\multicolumn{5}{l}{NER + Span Locating}\\[.1cm]
    INPUT:&\multicolumn{5}{l}{\texttt{PER}, \texttt{LOC}}\\
    INDEX:&0 & 1& 2 &3 &4 &5 &6& 7\\    
    INPUT:&Apple & CEO& Tim &Cook &sold &his &Texas& house\\    
     &$\uparrow\uparrow$ & & $\uparrow$ &$\uparrow $& & &$\uparrow\uparrow$& \\    
    TRANSFORM:& Start \& End & & Start & End & & & Start \& End & \\[.1cm]   
    OUTPUT:&\multicolumn{5}{l}{[$(0, 0, $~\texttt{ORG}$)$, $(2, 3, $~\texttt{PER}$)$, $(6, 6, $~\texttt{LOC}$)$]} \\
    \end{tabular}
\end{table}

Given a class label there are two typical ways to identify the span(s). First, given a input sentence with $\ell$ tokens, one way is to employ two $\ell$-class classifiers to predict the span boundaries (\ie, the start and end tokens)~\cite{Ding-EMNLP'22-ask}. The second way is to employ two binary classifiers, one to predict whether each token is a starting token or not, and the other to determine whether each token is the end token or not, with the obvious restriction that the beginning token must precede the end token~\cite{Li-ACL'20-unified}. The span locating transformation illustrated in the example shows that first the \texttt{PER} and \texttt{LOC} classes are identified either via a input or another model. Then the transformation seeks to identify the boundary tokens that begin and end the span that represents the class labels within the sentence.

\subsubsection{Span Generation}
\label{sec:spangen}
Text generation models have become popular especially with the rise in LLMs. These generation models provide another possibility in span-oriented transformations. The core idea of span generation is to transform the original token sequence into an expanded token sequence with span-tokens~\cite{Lewis-ACL'20-bart, Raffel-JMRL'20-t5}, similar to machine translation and other natural language generation tasks. In order to represent span labels in the token sequence, span generation  typically inserts distinct characteristics to indicate span labels including both span positions and span classes~\cite{DeCao-ICLR'21-GENRE, Lu-ACL'22-UIE, Fei-NIPS'22-LasUIE}. 

\begin{table}[h!]
    \centering
    \footnotesize
    \begin{tabular}{>{\raggedleft\arraybackslash}p{2.0cm} P{.5cm}P{.5cm}P{.5cm}P{.45cm}P{.45cm}P{.3cm}P{.5cm}P{.5cm}P{.3cm}P{.3cm}P{.5cm}P{.5cm}P{.5cm}P{.5cm}}
    TASK:&\multicolumn{7}{l}{NER + Span Generation}\\
    INDEX:&0 & 1& 2 &3 &4 &5 &6& 7& 8& 9& 10 & 11 & 12 & 13\\    
    INPUT:&Apple & CEO& Tim &Cook &sold &his &Texas& house & & & & & & \\    
    TRANSFORM:&\texttt{ORG-L} &Apple & \texttt{ORG-R} & CEO& \texttt{PER-L}& Tim &Cook& \texttt{PER-R} &sold &his &\texttt{LOC-L} &Texas& \texttt{LOC-R}& house \\[.1cm]
    OUTPUT:&\multicolumn{7}{l}{[$(0, 0, $~\texttt{ORG}$)$, $(2, 3, $~\texttt{PER}$)$, $(6, 6, $~\texttt{LOC}$)$]} \\
    \end{tabular}
\end{table}

Continuing the running example illustrated above, span generation might transform the input token sequence into a token sequence having spans represented by special span tokens like \texttt{[PER-L]} indicating the start of a person span or \texttt{LOC-R} indicating the end of a location. Span generation models are commonly designed as an autoregressive token generation task~\cite{DeCao-ICLR'21-GENRE, Lu-ACL'22-UIE, Fei-NIPS'22-LasUIE}. They take a token sequence and a generated token as input, and predict the subsequent token. These predictions essentially represent classifications from a predefined dictionary. During inference, the generation process continues iteratively until all the desired spans are generated. 


The span generation approach is widely utilized in many information extraction tasks. For example, GENRE employs an auto-regressive model to transform the entity disambiguation and entity linking tasks into a joint text/entity-name generation task~\cite{DeCao-ICLR'21-GENRE}. For the entity disambiguation task, target entities are selected with a conditional generation method based on the provided token sequence. As for the entity linking task, the span indices and their labels are together transformed into an augmentation of the original sentence. Specific to the NER task, the BartNER model transforms the token sequence into uniform index pointers~\cite{Yan-ACL'21-bertner}. And more recently, Universal Information Extraction (UIE) architectures have also been developed to transform different information extraction tasks (\eg, NER, EL, ED) into the same format through generative language modeling~\cite{Lu-ACL'22-UIE, Fei-NIPS'22-LasUIE}. UIE systems can extract shared features and joint correlations from training labels of different information extraction tasks. Furthermore, different structural signals across different information extraction tasks can be encoded into similar text allowing efficient and effective knowledge transfer from pre-trained models~\cite{DeCao-ICLR'21-GENRE, Ayoola-NAACL'22-refined}. 

\subsection{Transformations Trade Offs}

\setlength{\tabcolsep}{2pt}

\begin{table}[t]
\centering
\footnotesize
\caption{Trade-offs of Transformation Approaches}
{\renewcommand{\arraystretch}{1.1}
\begin{tabular}{@{}r ccc c ccc c ccc@{}}
\toprule
& Complexity & Class Distr. & \multicolumn{3}{c}{Features} & \multicolumn{2}{c}{Special Cases} &\\
\toprule
Transformation & \# of inst. & +/- & Token & Span & Span-Class & Nested & Discontinuous &\\
\midrule
Sequential Labeling  & $O(1)$ & 4/4 & \checkmark & \\
Token Prototype  & $O(1)$ & 4/4 & \checkmark \\ 
Token-pair Classification & $O(1)$ & 4/60 & \checkmark & & & \checkmark & \checkmark   \\
Span Classification & $O(1)$ & 3/27 & & \checkmark & \checkmark & \checkmark \\
Span Locating & $O(K)$ & 6/0 & \checkmark &  & \checkmark & \checkmark\\
Span Generation  & $O(\ell)$ & 6/8 & \checkmark &  & \checkmark &  \\

\bottomrule
\end{tabular}
}
\label{tab:tradeoff}
\vspace{.25cm}
\end{table}

As we alluded to in the previous section, different transformations have distinct trade-offs. Following Tab.~\ref{tab:tradeoff}, we present these considerations along four dimensions: (1) computational complexity, (2) the number of positive and negative class labels, (3) the features considered, and (4) applicability to nested and discontinuous spans. It is important to note that our discussion focuses on the general setup of these transformations without considering any specific design modifications. Although some models and methods may have specific designs tailored to address certain special cases, it is not our intention to delve into specific design strategies in the following discussion.

\subsubsection{Complexity}
Given an input document with $\ell$ tokens and $K$ different target span classes, different transformation approaches have different computational complexities. The main difference in complexity is the number of actual instances (\ie, \# of instances in Tab.~\ref{tab:tradeoff}). This count directly corresponds to the number of times the feed-forward process needs to be executed in order to generate an inference output. Most transformations require a single input, namely, the token sequence, and makes several span predictions. The span locating transformation needs to consider each provided span class as an individual instance and is therefore in $O(K)$. Like machine translation, the span generation transformation considers each token in the input sequence as an individual instance yielding $O(\ell)$.

\subsubsection{Positive and Negative Span Distributions}
Different transformation approaches produce a different number of targets, \ie, positive and negative spans, and therefore the choice of transformation has a significant impact on the label distribution and, as a result, the performance metrics. 

Again consider the running example illustrated above, which includes two named entities. In the sequence labelling transformation, four tokens including \texttt{Apple}, \texttt{Tim}, \texttt{Cook}, and \texttt{Texas} are positive instances representing span tokens; the other four tokens are negative instances representing non-span tokens. In token-pair classification, there are $\ell^2=64$ token pairs in total; of which two are positive samples. In span classification, with a window-length of five, there are a total of $30$ possible spans with length less or equal to $5$ from which only three represent positive spans. Finally, the span generation transformation produces three positive spans using six special tokens to annotate positive spans. Therefore the six special tokens are the positive instances and the original eight tokens are considered negative tokens. Clearly,  different transformation approaches yield substantially different class distributions, which by definition has a large impact on performance metrics. 

Noisy or incomplete labels are also differently impacted by the choice of transformation. Understanding these differences is important because many information extraction datasets have noisy training labels of $5\%$ or more~\cite{CrossWeigh-EMNLP'19-Han, NLL-EMNLP'21-Chen} even for the well-known CONLL03 NER dataset~\cite{Sang-NAACL'03-conll03}. 
In an interesting empirical study on the missing labels for information extraction tasks, Li \textit{et al.} considered as a special case where noise is only present in negative samples~\cite{NegSampling-ICLR'21-Shi}. They show that, during training, ignoring positive examples has small impact, but incorrectly labelling positive spans as being negative samples has significant impact on the final results. The same idea also applies for different transformations with different positive and negative span distributions.

\subsubsection{Features}
Different transformations use various models to produce spans, resulting in different abilities to encode different types of features. We focus on three main types of features: (1) token embeddings, (2) span embeddings, and (3) the span-class representation. 

Token-wise transformations such as sequential labeling, token prototype, and token-pair classification decompose span labels into token-wise classes, making token features easy to encode but precluding the consideration of span features. In span classification, the embeddings of span candidates can be obtained and then span classes can be assigned to the entire span. In contrast, span locating encodes the span class along with the original sentence as input, and results are obtained by locating the span boundaries using the span indices, meaning that the span embedding cannot be considered. Likewise, span generation cannot use span embeddings, but instead represents span classes as a sequence of tokens to generate.


\subsubsection{Nested and Discontinuous Spans}
Overlapping spans is an important complication for training, inferences, and evaluation. This occurs when a single token is made to belong to two or more different spans. Different transformations handle these cases in different ways. We categorize them into two distinct cases: nested and discontinuous.

Nested spans are situations where a token can belong to multiple spans simultaneously. Transformations which allow for nested spans include token-pair classification, span classification, and span locating. Token-pair considers nested cases into different start-end token pairs; span classification considers all the possible span candidates, which includes overlapping spans; and span locating has no restriction on where the beginning and ending indices of different classes may be placed. In contrast, sequential labeling and token prototype transformations decompose span labels into token-wise labels; as a result, a token can not belong to two classes simultaneously. Likewise, span generation requires the injection of special tokens to indicate span classes, and therefore can not provide nested spans either. 

Discontinuous cases, on the other hand, involve the formation of spans using non-adjacent tokens. In these cases, tokens that are not contiguous in the text can be grouped together to form a single span. Because the token-pair matrix provides the flexibility to link non-contiguous tokens, it is the only method that can effectively handle discontinuous cases.

\subsection{Training Strategies}

After a span transformation approach is chosen, the systems needs to be properly employed to be effective. In other words, different training strategies have to be considered for a working framework. From our perspective, we categorize the various standard training strategies into four broad classes: (\ref{sec:feattune}) Feature tuning, which selects and engineers the most relevant features for a particular information extraction task; (\ref{sec:modeltune}) Model tuning, which refers to the process of optimizing the parameters of a machine learning model; (\ref{sec:prompttune}) Prompt tuning, which is the relatively new task of fine-tuning the prompts fed to LLMs to achieve more accurate results; and (\ref{sec:incontext}) In-Context learning, which involved training models in specific context information, for example, on specific datasets or niche tasks. 

\subsubsection{Feature Tuning}
\label{sec:feattune}
One of the outputs of pre-trained (large) language models are informative and well-trained embeddings. These embeddings almost always represent a span---sometimes as short as a word and sometimes as long as a whole sentence or paragraph---and are effective features that can be used for information extraction tasks. Prior to the rise of LLMs, most features used in information extraction tasks came from linguistic cues such as part-of-speech (POS) tags~\cite{Rijkhoff-Language_and_Linguistics_Compass'07-word_classes}, word stems and lemmas~\cite{Bird-ACL_workshop'04-nltk}, and syntactic parsers~\cite{Pickering-Handbook_of_psycholinguistics'06-syntactic_parsing}; as well as statistical learning approaches like word frequency counts~\cite{Aizawa-IPM'03-tf_idf}, word co-occurrence analysis~\cite{Mikolov-NIPS'13-word2vec}, and semantic analysis~\cite{Landauer-Discourse_processes'98-semantic_analysis}. Although these methods have their own strengths and weaknesses, they are generally less effective than LLM-based embeddings in capturing the complex relationships between words in a language\cite{Mikolov-NIPS'13-word2vec, Devlin-NAACL'19-bert, Brown-NIPS'20-gpt3, Lewis-ACL'20-bart, Raffel-JMRL'20-t5}. Feature turning, therefore, refers to the numerous strategies that have been developed to learn ever-more creative and interesting features for spans. Equipped with these pre-trained features, spans can be clustered or classified or labeled to solve any number of information extraction tasks.

\subsubsection{Model Tuning}
\label{sec:modeltune}
Sometimes, the pre-trained features from an LLM do not align well with the task that is being asked of the system. This misalignment will degrade the system's performance. In these cases it is common for LLMs to undergo a fine-tuning process, which adapts the pre-trained model parameters, including span features, to the specific task. Previous studies have shown that fine-tuning the model outperforms feature engineering with similar settings~\cite{Devlin-NAACL'19-bert, Lewis-ACL'20-bart}, but can be prone to catastrophic forgetting~\cite{Neubig-Computing_Survey'23-prompt} and other maladies; see the survey by Li \textit{et al.}~\cite{li2021pretrained} for details. One major problem is that model tuning requires that the model be loaded into memory and trained, which, for even medium-sized LLMs, is a non-trivial task.

\subsubsection{Prompt Tuning} 
\label{sec:prompttune}
The rise of ChatGPT and other proprietary LLMs has spawned an entirely new kind of NLP task called prompt-tuning. In this case, instead of using or training span embeddings, prompt-tuning is the task of devising clever ways to query the LLM. The advantages of prompt tuning are clear. Because there is no need to extract, tune, or train any model or features, it is relatively easy to use the system. The task instead becomes finding the best prompts to feed to the LLM so that it returns the answers you seek. Another often overlooked advantage of prompt tuning is that by simply querying the system, it does not change. As a result any prompts or other rules that are learned can be maintained. 

\begin{table}[h!]
    \centering
    \footnotesize
    \vspace{0.5cm}
    \begin{tabular}{>{\raggedleft\arraybackslash}p{2cm} p{4.2in}}
    TASK:& NER \\
    INPUT: & Apple CEO Tim Cook sold his Texas house. 
    \\
    Prompt INPUT: & Apple CEO Tim Cook sold his Texas house. Tim Cook is a [MASK] \\
    Prompt Output: & [MASK] $\rightarrow$ technology executive \\
    OUTPUT:& $[(2, 3, $~\texttt{PER}$)$] \\
    \end{tabular}
\end{table}

There are two types of prompt tuning: hard prompt tuning and soft prompt tuning. In hard prompt tuning, a handcrafted prompt is used to glean results from the system. Conversely, in soft prompt tuning, the prompt itself can be trained. This means that an additional NLP model is trained to predict an adaptable prompt based on some input and labels. During the inference process, the prompt model initially generates a prompt, which is then concatenated with a test sample and fed into the LLM to obtain the final prediction; \ie, soft prompt tuning is a model generating input to feed to another model.

\subsubsection{In-context Learning} 
\label{sec:incontext}
Finally, In-context learning uses language models directly without any extra training process. This is accomplished by also injecting a few training examples along with corresponding labels into the prompts. The idea is that language models are able to see the mapping function between example input data and their corresponding labels, and then they can subsequently infer that same correspondence on unseen input data for label prediction.

\begin{table}[h!]
\begin{tabular}{p{\linewidth}}
\quad An example of in-context learning on the NER task might resemble something like this:\\[1em]
\end{tabular}
    \centering
    \footnotesize
    \begin{tabular}{>{\raggedleft\arraybackslash}p{2cm} p{4.2in}}
    TASK:& NER \\
    INPUT: & Apple CEO Tim Cook sold his Texas house. 
    \\
    Prompt INPUT: &Satya Nadella says that Microsoft products will soon connect to OpenAI.\\
    &Satya Nadella is a PER \\
    & Apple CEO Tim Cook sold his Texas house. Tim Cook is a [MASK] \\
    Prompt Output: & [MASK] is a PER \\
    OUTPUT:& [$(2, 3, $~\texttt{PER}$)$] \\
    \end{tabular}
\end{table}


With in-context learning a system can achieve robust capabilities with little cost. Another instance of this kind of learning is the chain-of-source approach, wherein a question and its corresponding answer are broken down into a series of sub-problems\cite{Neubig-Computing_Survey'23-prompt}. By addressing these sub-questions in a sequential manner, the system is able to arrive at more-comprehensive and nuanced solutions. 

\section{Discussion}

Over the past millennia text has been made \textit{by} humans \textit{for} humans. The recent and broad digitization of human-generated text has served to propel AI systems and tasks like information extraction. When humans are tasked to perform information extraction, they do so---with relative ease---by first understanding the concepts and definitions of the labels in their context, even on unseen classes or in unknown languages. In contrast, we often find that AI systems on few-shot and zero-shot scenarios still perform much worse than humans~\cite{Ding-arxiv'21-prompt_et, raiman-AAAI'22-deeptype2}. This gap in performance is due to the AI system's inability to reason about the relationships between the context of the input and the context of the class label. Ongoing work is in this area has aims to properly encode these contexts. For example, previous work in encoding entity descriptions as search query targets has shown some ability to retrieve relevant entity candidates~\cite{Wu-EMNLP'20-scalable, logeswaran-ACL'19-zero}. However, these relationships are nuanced; this research gap has not been fully explored and a wide gap remains. 


Most modern information extraction systems are built atop token-based pre-trained models. They work by tying together token features to identify spans. However, as pre-trained LLMs like GPT-4 and beyond grow ever stronger, they also become less accessible for specific use in AI tasks like information extraction. Therefore, we expect the use of in-context learning and further development of context-based prompts will increase. Ongoing work in retrieve-to-augment and generate-to-augment strategies have potential to enhance performance in many information extraction tasks as they have already been successful in open-domain question answering (ODQA) tasks~\cite{Yu-ICLR'23-generate}. 


In summary, the goal of this survey is to challenge the reader to reconsider information extraction to be the task of finding spans in text that represent some class-label. By recasting the myriad traditional information extraction tasks into this new orientation, they can all be viewed as analogs of one another. Having a unified view of this critical task is important as AI undergoes a shift due to the advent of LLMs and other groundbreaking AI models. To that end, we have summarized the various information extraction tasks, their evaluations, and their models, and we have endeavored to provide guidance how when and how to use each. We conclude with a challenge to integrate information extraction into LLMs so as to provide these sometimes rambling and hallucinating AI systems a bit of context and evidence; and, conversely, to better use the machinery of these LLMs to help tackle the information extraction task directly.

\section*{Acknowledgements}
This project was funded by DARPA under contracts HR001121C0168 and HR00112290106.




\end{document}

%% file: table/task.tex
{
\begin{tabular}{@{}l ccc c ccc c ccc@{}}
\toprule
\multicolumn{4}{c}{\textbf{Apple} CEO \textbf{Tim Cook} sold his \textbf{Texas} house.} \\
\midrule 
Task Name & Span Req. & Span Class & Span Class Example \\
\midrule
Entity Disambiguation (ED) & \checkmark & Entity & wiki/Tim\_Cook, wiki/Texas \\
Entity Linking (EL) &  & Entity & \url{wiki/Tim_Cook}, \url{wiki/Texas} \\
Entity Typing (ET) & \checkmark & Fine-grained Type  & \textsf{Businessman}, \textsf{State}    \\
Named Entity Recognition (NER) &  & Coarse-grained Type  & \texttt{PER}, \texttt{LOC}    \\
Attribute Value Extraction (AVE) &  & Attributes & \textsf{CEO}: \texttt{Tim Cook} \\
Machine Reading Comprehension (MRC) &  & Reading Question & \texttt{Tim Cook} \\

\bottomrule
\end{tabular}
}